\documentclass[a4paper,twoside]{article}

\usepackage{epsfig}
\usepackage{subcaption}
\usepackage{calc}
\usepackage{amssymb}
\usepackage{amstext}
\usepackage{amsmath}
\usepackage{amsthm}
\usepackage{multicol}
\usepackage{pslatex}
\usepackage{apalike}
\usepackage{algorithm2e}
\usepackage[bottom]{footmisc}

\usepackage{bm}
\usepackage[subrefformat=parens]{subcaption}
\usepackage{SCITEPRESS}

\newtheorem{cond}{Condition}[section]

\begin{document}

\title{Neural Bradley-Terry Rating: Quantifying Properties from Comparisons}

\author{\authorname{Satoru Fujii}
\affiliation{Kyoto University}
\email{fujii.satoru.75c@st.kyoto-u.ac.jp}
}


\abstract{Many properties in the real world don't have metrics and can't be numerically observed, making them difficult to learn. To deal with this challenging problem, prior works have primarily focused on estimating those properties by using graded human scores as the target label in the training. Meanwhile, rating algorithms based on the Bradley-Terry model are extensively studied to evaluate the competitiveness of players based on their match history. In this paper, we introduce the Neural Bradley-Terry Rating (NBTR), a novel machine learning framework designed to quantify and evaluate properties of unknown items. Our method seamlessly integrates the Bradley-Terry model into the neural network structure. Moreover, we generalize this architecture further to asymmetric environments with unfairness, a condition more commonly encountered in real-world settings. Through experimental analysis, we demonstrate that NBTR successfully learns to quantify and estimate desired properties.}

\onecolumn \maketitle \normalsize \setcounter{footnote}{0} \vfill

\section{\uppercase{Introduction}}

There are multitudes of properties humans can recognize. For some of them, we have well-defined metrics: grams for weight, decibels for loudness, or sometimes it's just a number for counting. However, most properties can't be simply measured, and some can't be even directly observed: We don't have metrics for the ``attractiveness'' of merchandise, nor can we directly observe the ``strength'' of a deck in card games. Our goal is to quantify those properties and obtain an estimator of it.

One way to tackle this problem is to conduct a survey and ask people to rate those properties on a scale of 1 to 5, then use those scores as the target label of supervised learning. However, those values don't have any meanings other than just being high or low, making them rather superficial metrics.

On the other hand, many rating algorithms based on Bradley-Terry Model \cite{bradley1952rank} have been studied to estimate the strength of players in a competitive environment. These methods allow us to quantify the strength of players based on their match histories: winning against other players makes their estimated rating higher, more so if the opponent's rating is high. However, those methods can be used only on known items such as human players, and thus can't predict properties of unknown ones which aren't present in the comparison dataset.

In this paper, we introduce \textbf{the Neural Bradley-Terry Rating (NBTR)}, an ML framework that seamlessly integrates traditional rating based on the Bradley-Terry model into neural networks. This network learns the quantification of properties based on the results of comparisons. For example, our framework can be used to obtain an estimator of:
\begin{itemize}
     \item Appeal of the price and the description or package illustration of products, by learning what was bought on e-commerce services or vending machines.
	\item Attractiveness of the name and the thumbnail image or preview text, by learning what was clicked on the internet search engine or online video platform.
	\item Strength of a deck in card games, by learning match histories.
	\item Beauty of a painting or palatability of a dish, by learning a result of a human survey regarding their preference.
\end{itemize}

We also introduce a generalization that discounts the unfairness of the comparison to estimate ratings. For instance, on internet search engine or e-commerce platform, we usually prioritize the ones shown on the upper side, and we rarely keep scrolling down and going next pages, resulting in unfair comparisons. Our generalized NBTR architecture is designed to be applied to those asymmetric situations.

In some cases, comparison datasets can be obtained without any surveys, like the first 3 examples where we can automatically record users' choices. While learning some properties still requires a human survey, it's easier to answer ``which do you prefer'' questions than scoring them in a particular grade. This makes the survey more reliable and precise. Furthermore, since our method is based on the Bradley-Terry model, the rating it yields is a scalar representation of competitiveness, which can be used to estimate probabilities of winning or being selected over others.

We empirically show that our method is capable of obtaining good quantification and estimator of the target property. We also show the validity of our network structure by comparing several possible variants in our experiment.

\section{\uppercase{Background}}

\subsection{Bradley-Terry Model}

\textbf{\textit{Bradley-Terry score}} $\pi_i$ is a scalar which represents the competitiveness of player $i$. In this paper, we will generalize this as the strength of a certain property of item $i$. Let $W_{ij}$ be the winning probability of player $i$ against player $j$. Bradley-Terry model \cite{bradley1952rank} assumes:
\[
W_{ij} = \frac{\pi_i}{\pi_i + \pi_j}
\]

Note that $(\pi_1, \dots ,\pi_N)$ is scale-free. In other words, $(\pi_1, \dots ,\pi_N)$ is the same as $(c\pi_1, \dots , c\pi_N)$ for any constant $c$ in terms of winning probabilities. We simply can fix the average to uniquely determine $(\pi_1, \dots ,\pi_N)$.

Given this model, we want to calculate the Maximum Likelihood Estimation (MLE) of $(\pi_1, \dots ,\pi_N)$ from a match history among players $1, \cdots , N$. However, there are situations where calculating MLE is infeasible. For example, if player $i$ won at least once and never lost a single match, MLE of $\pi_i$ will diverge to infinity. To calculate MLE, the following condition should be met \cite{doi:10.1080/00029890.1957.11989117}:

\begin{cond} \label{condi}
In every possible partition of the players into two nonempty subsets, some player in the second set beats some player in the first set at least once.
\end{cond}

Numerical approaches such as Minorization-Maximization (MM) algorithm \cite{hunter2004mm} are commonly used to approximately calculate these MLE scores.

\subsection{Generalizations of Bradley-Terry Model}
The Bradley-Terry model can be generalized for settings where matches are held among more than 2 players by assuming the winning rate of player $i$ in a match among player $1, \dots , M$ to be:
\begin{equation} \label{genbt}
\frac{\pi_i}{\sum_{k = 1}^M \pi_k}
\end{equation}
This can be also applied to a situation where the ranks of players are decided rather than just a single winner, by regarding the 2nd place to be the winner of the competition of remaining $M-1$ players.

For asymmetric environments where players compete in unfair settings, a variant of the Bradley-Terry model also has been proposed. It assumes
\begin{equation} \label{home}
W_{ij} = \frac{\eta \pi_i}{\eta \pi_i + \pi_j}
\end{equation}
where $\eta > 0$ is the strength of player $i$'s advantage.

Calculating MLE under those generalized models is also possible by using MM algorithms \cite{hunter2004mm}.

\subsection{Elo Rating}
Elo Rating \cite{elo1978rating} is an incremental approach to estimate strength. It defines \textbf{\textit{rating}} of player $i$ as
\begin{equation} \label{elo}
R_i := \alpha \log_{10} \pi_i + \beta
\end{equation}
usually with $\alpha = 400, \beta = 1500, E[\pi_{i}] = 1$. Using a logarithm for rating prevents the value from scaling too much.

In Elo Rating, every player starts with a rating of $\beta$. When player $i$ wins $w$ times among $g$ matches against player $j$, rating of player $i$ will be refreshed as
\[
R'_i \gets R_i + k (w - g W_{ij})
\]
where $k$ is a learning rate. $W_{ij}$ is calculated from $R_i$ and $R_j$.

Microsoft Research has proposed TrueSkill \cite{herbrich2006trueskill}, which combined the probability graph model with Elo Rating and achieved increased accuracy and convergence speed.

\subsection{NN Modifications and Expansions}

In some cases, we want to use shared weights for certain parameters of NN. This idea is introduced in the context of natural language processing \cite{inan2016tying}.

Skip connection \cite{he2016deep} is a method that directly adds the output of some layer to another, which was introduced and proved to be successful in image recognition.

Using NN structure to transform the data into a vector with lower dimensionality is a commonly accepted idea. Autoencoder \cite{hinton2006reducing} is arguably the first one of those methods, which learns dimensionality reduction by using the input itself as a learning target with a small hidden layer.

\section{\uppercase{Related Work}}

\subsection{User Score Prediction}
There has been extensive research in the area of user score prediction such as 1-5 graded ratings. Collaborative Filtering \cite{Jalili2018EvaluatingCF} is an actively studied field, which aims to predict user ratings based on past ratings. The problems of using user rating as ground truth are also recognized in this field. Deep Rating and Review Neural Network (DRRNN) \cite{9446537} uses review text as an additional target for back-propagation to mitigate these problems.

\subsection{Bradley-Terry Model and NN}
There has been a small number of research that combines the Bradley-Terry model with the neural network. 

NN Go rating model \cite{gorating} has been proposed to estimate ratings of Go players present in match history, combined with estimated intermediate winning rate during a match and history decay of rating, and outperformed traditional rating algorithms including Elo and TrueSkill. Both our objectives and the network architecture are fundamentally distinct.

A previous research \cite{li2021neural} used a neural network to predict image beauty scores. They calculated the MLE of Bradley-Terry scores of images by MM algorithm, then used winning probabilities based on those scores as the target labels in the training of NN. They focused on their specific image beauty task with their controlled dataset, which resulted in important differences between our approaches and theirs. Our method doesn't require any outside pre-calculation of Bradley-Terry scores, which enables online learning and simplifies the implementation. Furthermore, in a data-collecting environment that isn't statistically controlled, Condition \ref{condi} might not be held, which makes the calculation of MLE impossible. Even if that's not the case, since Bradley-Terry scores for an item with a small number of matches would be less reliable, their method will be unable to properly weigh the loss during the training. We also generalized the architecture further for asymmetric environments, where calculating the MLE score is infeasible without using some simplification like (\ref{home}).

There also has been a work claiming to involve Bradley-Terry artificial neural network model \cite{menke2008bradley}. It introduced a single-layer network with 2 inputs fixed to -1 and 1 and was meant to be an iterative algorithm to obtain individual ratings, similar to Elo rating. Their structure is hardly a ``neural network model'' as we call it today, and our approach is simply different from theirs.

\section{\uppercase{Proposed Approach}}
\subsection{Symmetric Setting}
In this section, we describe the architecture of NBTR. The goal of NBTR is to estimate the rating, that is, the quantification of a certain property of unknown items that haven't appeared in comparison histories. Formally, we aim to obtain a NN we call \textbf{\textit{rating estimator}} $E: \bm{x}_i \mapsto R_i$, where scalar $R_i$ is the rating of item $i$.

The training of NBTR uses a dataset where each entry contains explanatory variables of $M(\geq 2)$ items (not necessarily human competitors), and the results of comparison $\bm{y}$, like which athlete won the match, or which thumbnail was clicked. $\bm{y}$ is a $M$-dimensional one-hot vector which represents what item won the comparison. Formally, the shape of each entry of the training data is $(\bm{x}_1, \cdots, \bm{x}_M, \bm{y})$, where $\bm{x}_i$ is the vector representation of item $i$.

\begin{figure}
 \centering
 \includegraphics[keepaspectratio, scale=0.32, bb=0 0 849 464]
      {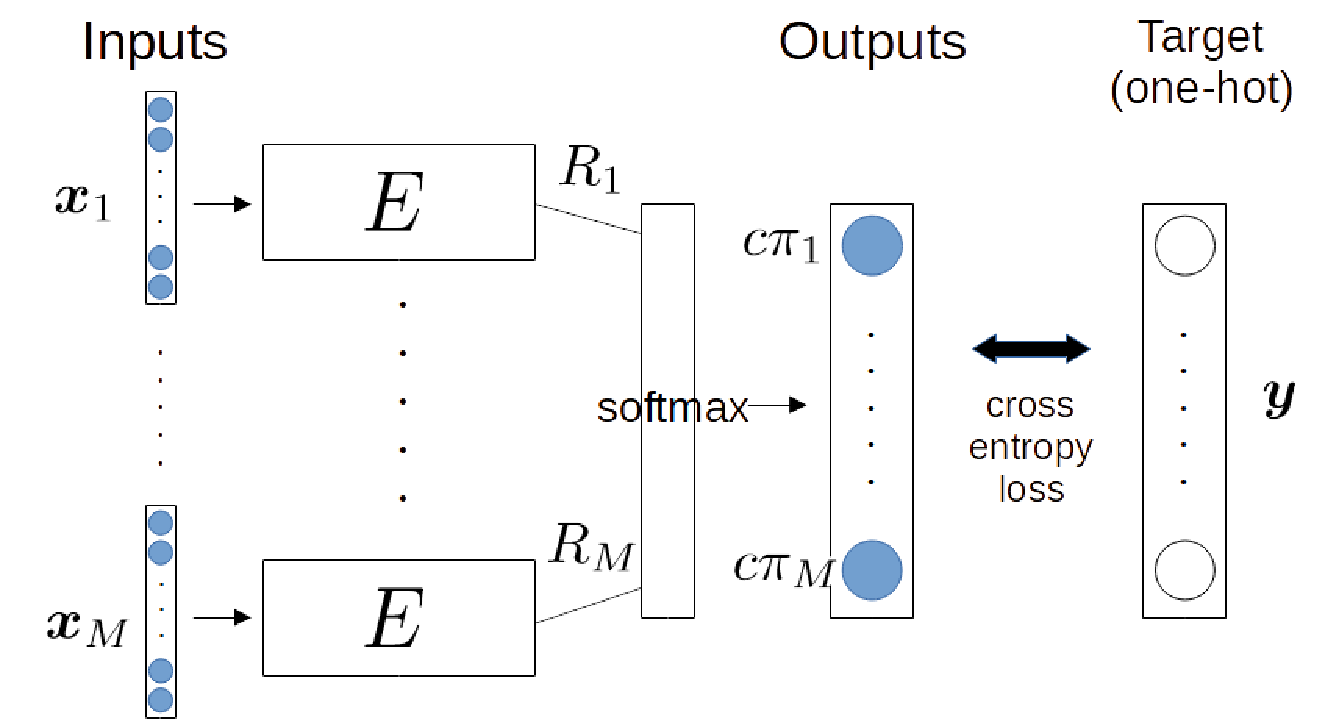}
 \caption{The structure of NBTR for symmetric environments} \label{structure_symmetric}
\end{figure}

To obtain $E$, we connect the outputs of $M$ rating estimators with shared weight via softmax function $(R_1, \cdots R_M) \mapsto (c\pi_1, \cdots, c\pi_M)$ where
\[
\pi_i = e^{R_i}, \ c = \frac{1}{\sum_{k=1}^M \pi_i}
\]
then use $\bm{y}$ as a target label with cross-entropy loss. The output $(c\pi_1, \cdots, c\pi_M)$ is exactly the same as (\ref{genbt}). This structure is shown in Figure \ref{structure_symmetric}.

Since NN learns its parameters to minimize the difference between $(c\pi_1, \cdots, c\pi_M)$ and actual comparison result $\bm{y}$, $\pi_i$ will be the equivalent of Bradley-Terry scores $\pi_i$ after enough training. For this reason, we used the same notations for them. Since $R_i = \log \pi_i$, $R_i$ should be the equivalent of Elo Rating values, where $\alpha = 1, \ \beta = \log 10$ with a certain scale of $\pi_i$ in (\ref{elo}). Weight sharing forces rating estimators to calculate the ratings of each item in the same manner, allowing us to obtain a single rating estimator.

\subsection{Asymmetric Setting}
In many real-world environments, comparisons are not made under fair conditions. However, we still want to use NBTR in these asymmetric environments to output ratings, whose value can be used to estimate the results of comparisons as if items would be compared on a fair condition.

\begin{figure}
 \centering
 \includegraphics[keepaspectratio, scale=0.27, bb=0 0 1052 485]
      {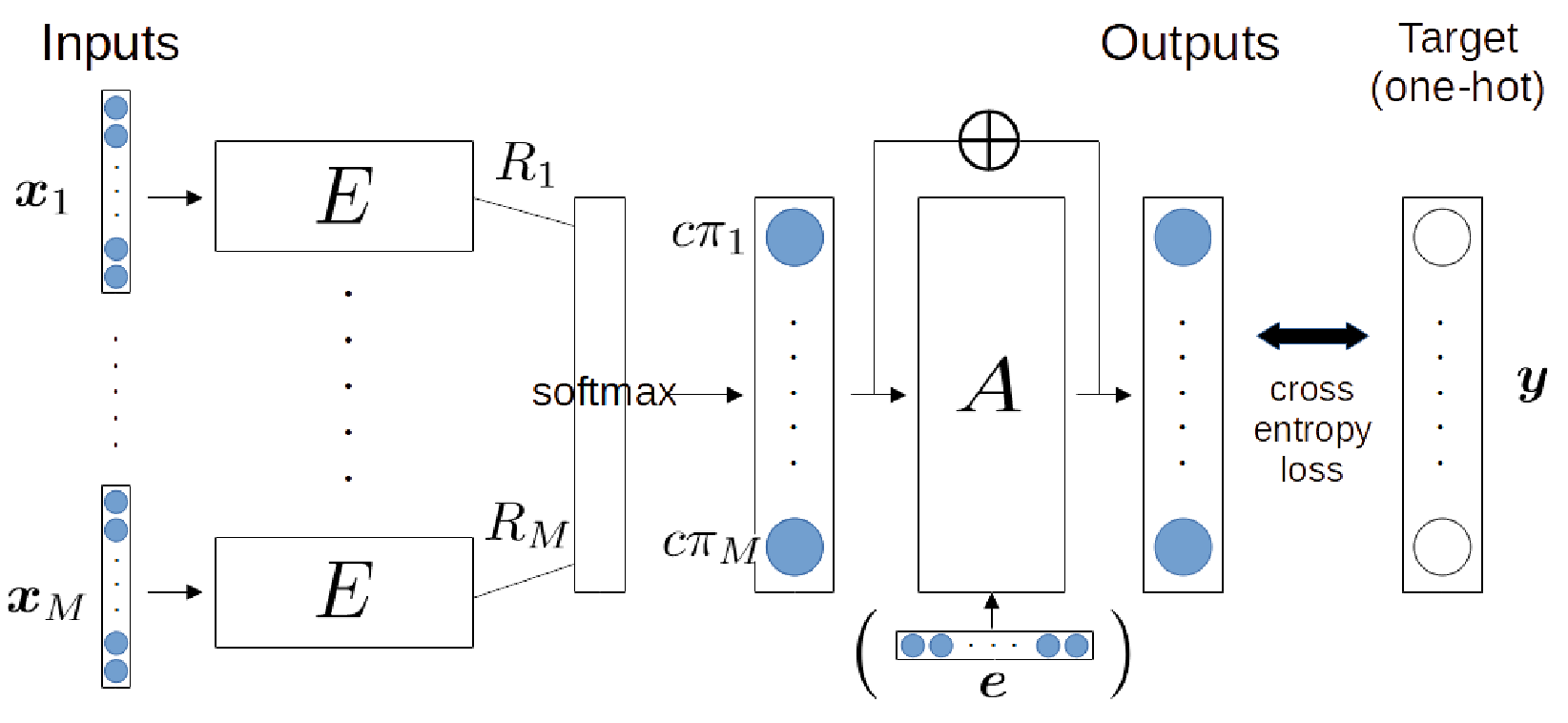}
 \caption{The structure of NBTR for asymmetric environments} \label{structure_asymmetric}
\end{figure}

To achieve this, we insert a NN $A$ we call \textbf{\textit{advantage adjuster}}, which takes $(c\pi_1, \cdots, c\pi_M)$ as an input, and aims to return the probability distributions of each item winning the comparison, based on learned unfairness of environment. We also use skip connection around it to prevent the situation where the rating estimator deviates from the intended rating settings due to the degree of freedom. For example, rating estimators may learn to output $-R_i$ instead of $R_i$ and advantage adjuster will still be able to adapt to it and properly predict the outcome. Skip connection is expected to solve this problem since $A$ would learn to do nothing if the environment turned out to be fair, and otherwise change its parameters to somewhere around there. This structure is shown in Figure \ref{structure_asymmetric}.

As a benefit of using NN, advantage adjuster should be capable of dealing not only with simple advantages such as (\ref{home}) but also with relations between positions. For instance, it should adapt to situations where people tend to overrate an item sandwiched between less appealing items.

When external factors are believed to affect the unfairness of the environment, it is possible to add an \textbf{\textit{environment vector}} $\bm{e}$ as an additional input to advantage adjuster. For example, home-court advantage might be stronger on a sunny day due to more spectators. In this case, the advantage adjuster should receive information about the weather as an additional input.

\section{\uppercase{Experiments}}

\subsection{Experiment for Symmetric Setting}
Firstly, to measure the performance of our method in a symmetric setting, we customized MNIST \cite{lecun1998gradient} dataset in a way that each entry has 2 images of handwritten letters and the result of the comparison, which is simply determined by which number is higher (random for ties).

For $E$, we used a simple NN with 2 hidden layers consisting of 512 nodes each. We used ReLU for the activation function and Adam \cite{kingma2014adam} for the optimizer. We transformed 60000 and 10000 entries of the original data into the training and test data with the same number of entries. Each image on the original data was compared 2 times in our transformation: with the image above and below. The number of epochs was 5. We tried not to optimize network parameters as that is not the goal of this paper.

\begin{figure*}
\begin{tabular}{cc}
\begin{minipage}[b]{0.48\linewidth}
 \centering
 \includegraphics[keepaspectratio, scale=0.76, bb=0 0 382 376]
      {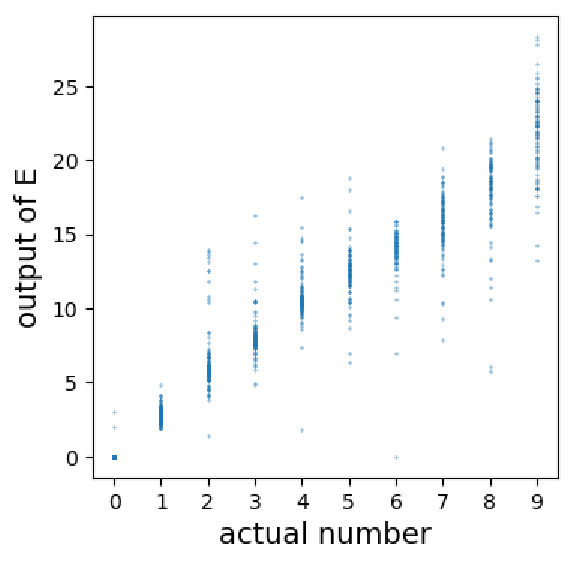}
 \subcaption{Symmetric} \label{exp_mnist_sym}
\end{minipage} &
\begin{minipage}[b]{0.48\linewidth}
 \centering
 \includegraphics[keepaspectratio, scale=0.76, bb=0 0 382 376]
      {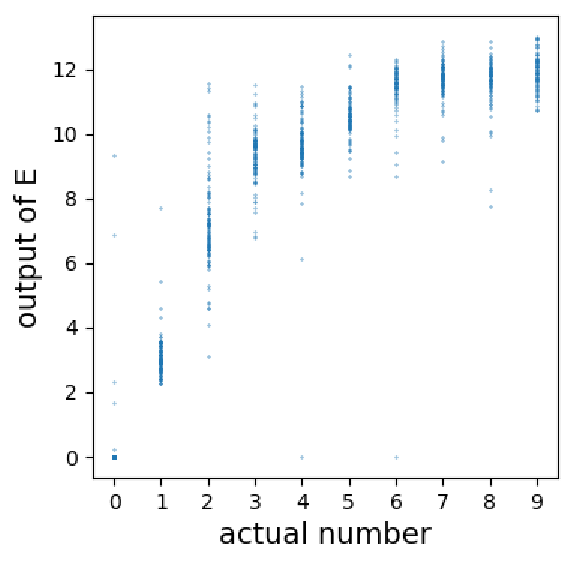}
 \subcaption{Asymmetric without $A$} \label{exp_mnist_asym_noadj}
\end{minipage} \\
\begin{minipage}[b]{0.48\linewidth}
 \centering
 \includegraphics[keepaspectratio, scale=0.76, bb=0 0 382 376]
      {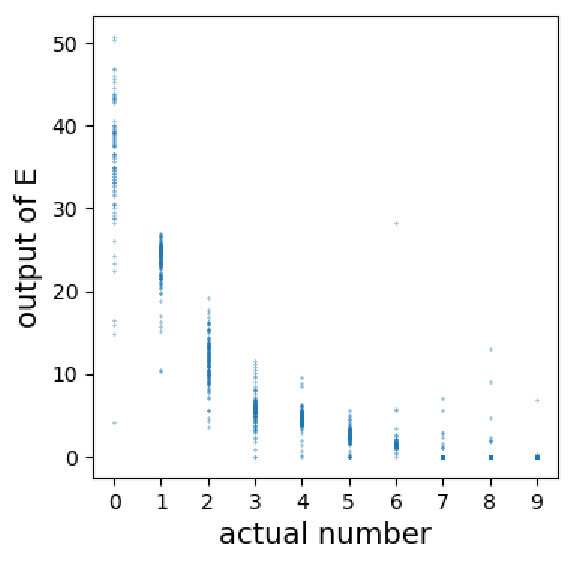}
 \subcaption{Asymmetric with $A$ without $\oplus$} \label{exp_mnist_asym_noskip}
\end{minipage} &
\begin{minipage}[b]{0.48\linewidth}
 \centering
 \includegraphics[keepaspectratio, scale=0.76, bb=0 0 382 376]
      {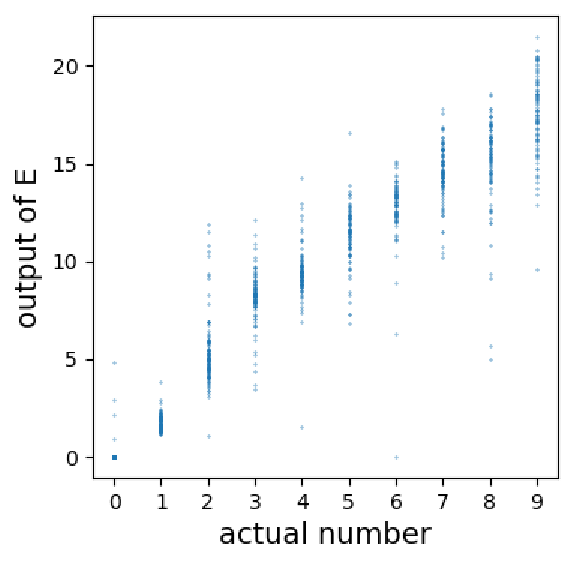}
 \subcaption{Asymmetric with $A$ and $\oplus$} \label{exp_mnist_asym}
\end{minipage}
\end{tabular}
\caption{The results of MNIST expetiments}
\end{figure*}

Figure \ref{exp_mnist_sym} is a scatter chart of the first 1000 entries in test data with the actual number of the image and the output of rating estimator $R_i$. This shows that our method successfully learned the quantification of ``number'' only from the results of the comparisons, despite the fact that this setting does not align well with the assumption of the Bradley-Terry model.

In a world where we know all actual numbers of the image, the ratio of $\pi_i : \pi_j$ would be $1 : \infty$ when the actual number of image $i$ lower than image $j$, since there's no chance $i$ would win against $j$. However, in our setting, the fuzziness of letters prevents this from happening. For example, the average value of $R_i$ on images of numbers 1 and 2 were 2.83 and 6.41 as shown in Table \ref{table}, which make $\pi_i$ 16.86 and 610.37, leading to 97.31\% win rate of 2 against 1. This would make sense considering the existence of a handwritten letter that looks like 1 while it's actually 7 or 9 with an extremely small curve on top.

\begin{table*}[tbp]
\begin{center}
{\tabcolsep = 0.65cm
\begin{tabular}{l|llll}
\hline
\multicolumn{1}{c|}{}  & \multicolumn{1}{c}{Symmetric} & \multicolumn{1}{c}{Asym wo/ A} & \multicolumn{1}{c}{Asym with A wo/ $\oplus$} & \multicolumn{1}{c}{Asym with A and $\oplus$} \\ \hline \hline
\multicolumn{1}{c|}{0} & 0.06 $\pm$ 0.39 & 0.24 $\pm$ 1.27    & 35.71 $\pm$ 7.73   & 0.13 $\pm$ 0.65 \\
\multicolumn{1}{c|}{1} & 2.83 $\pm$ 0.51 & 3.07 $\pm$ 0.61    & 23.46 $\pm$ 2.71   & 1.71 $\pm$ 0.38 \\
\multicolumn{1}{c|}{2} & 6.41 $\pm$ 2.18 & 7.25 $\pm$ 1.47    & 11.68 $\pm$ 2.78   & 5.31 $\pm$ 1.69 \\
\multicolumn{1}{c|}{3} & 8.20 $\pm$ 1.60 & 9.35 $\pm$ 0.80    & 5.7 $\pm$ 2.02     & 8.17 $\pm$ 1.31 \\
4                      & 10.72 $\pm$ 1.64 & 9.60 $\pm$ 1.17   & 4.61 $\pm$ 1.27    & 9.39 $\pm$ 1.30 \\
5                      & 12.40 $\pm$ 1.82 & 10.53 $\pm$ 0.62  & 2.48 $\pm$ 1.07    & 11.33 $\pm$ 1.51 \\
6                      & 13.73 $\pm$ 1.99 & 11.31 $\pm$ 1.37  & 1.9 $\pm$ 2.94     & 12.71 $\pm$ 1.84 \\
7                      & 15.69 $\pm$ 2.04 & 11.69 $\pm$ 0.55  & 0.26 $\pm$ 1.01    & 14.57 $\pm$ 1.40 \\
8                      & 17.71 $\pm$ 2.76 & 11.61 $\pm$ 0.75  & 0.39 $\pm$ 1.76    & 15.08 $\pm$ 2.31 \\
9                      & 21.85 $\pm$ 2.78 & 11.92 $\pm$ 0.54  & 0.08 $\pm$ 0.70    & 17.51 $\pm$ 2.05 \\ \hline                                                 
\end{tabular}
}
\caption{Average and standard deviation of $E$'s outputs} \label{table}
\end{center}
\end{table*}

\subsection{Experiment for Asymmetric Setting}
Secondly, to measure the validity of our network structure in asymmetric settings, we again customized the MNIST dataset mostly in the same way, except that this time the result labels were determined by the comparison of (1.4 $\times$ the number of the left image + 0.1) with the number of the right image. We used a simple 2-nodes linear layer for $A$, and the rest of the settings were the same as the symmetric setting.

We compared three network structures in this experiment: one without the advantage adjuster, one with the advantage adjuster but without skip connection, and the one with both, which we propose. They achieved 82.8\%, 93.7\%, and 94.6\% accuracy on the test dataset during the training, and the result of rating estimations are shown in Figure \ref{exp_mnist_asym_noadj}, \ref{exp_mnist_asym_noskip}, \ref{exp_mnist_asym} respectively. Note that due to the scale-free nature of Bradley-Terry scores, we will focus on the shape of the curve rather than the scale of the output values.

As shown in Figure \ref{exp_mnist_asym_noadj}, without the advantage adjuster, rating estimations were distorted by environmental unfairness. Although the structure without skip connection achieved almost the same accuracy as the full structure during the training, rating estimator failed to learn desired quantification as shown in Figure \ref{exp_mnist_asym_noskip}. This kind of deviation is likely to happen due to the degree of freedom as we discussed. Our proposed structure was able to denoise the unfairness of the environment and obtained a similar curve of rating estimation to the one in a symmetric setting, as shown in Figure \ref{exp_mnist_asym}.

\subsection{Experiment on Pok\'{e}mon Dataset}

Thirdly, we show a small example of the actual usage of our framework, closer to what we envision. We used Weedle's Cave \cite{pokemon} dataset, which contains 50000 match results between 800 Pok\'{e}mons, generated by a custom algorithm that omits some mechanics of the game. As explanatory variables, we used 6 base stats and an 18-dimensional 0-1 vector which represents whether a Pok\'{e}mon has a certain type or not. We separated $1/4$ Pok\'{e}mons for the test purpose, and only used matches that don't involve those Pok\'{e}mons for the training of NBTR (20\% of them were used for validation). Since Pok\'{e}mon battle is a simultaneous game, we used symmetric NBTR architecture. For $E$, We used a simple NN with 2 hidden layers consisting of 64 nodes each. The rest of the settings are the same as the MNIST experiments.

\begin{figure}
 \centering
 \includegraphics[keepaspectratio, scale=0.55, bb=0 0 386 375]
      {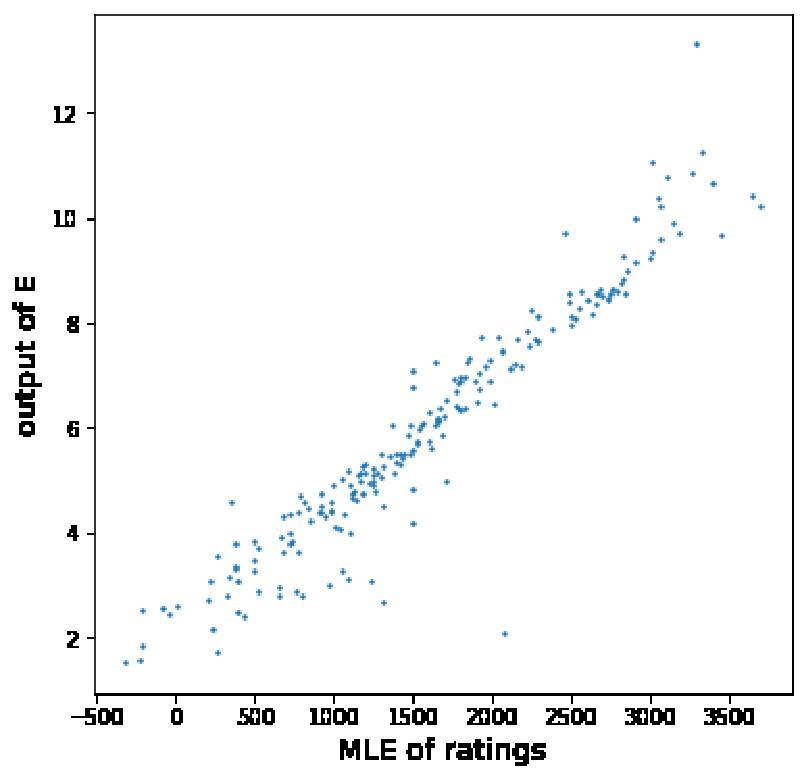}
 \caption{The outputs of $E$ and MLE ratings of test Pok\'{e}mons} \label{poke_exp_bt}
\end{figure}

We compared the outputs of $E$ and MLE ratings of test Pok\'{e}mons (not included in the training dataset). We calculated MLE scores from all match results using choix \cite{choix}, then transformed them in the same manner as Elo rating. As shown in Figure \ref{poke_exp_bt}, NBTR could estimate the proper ratings of unseen Pok\'{e}mons with a 0.95 correlation coefficient. Table \ref{poke_hyp} shows the estimated strength of imaginary Pok\'{e}mons or real Pok\'{e}mons not present in the dataset, obtained by the same $E$ as a showcase of the potential of our method.

\begin{table*}[tbp]
\begin{center}
{\tabcolsep = 0.4cm
\begin{tabular}{lllllll|l}
\hline
HP  & Attack & Defence & Sp. Atk & Sp. Def & Speed & Type           & Output of E \\ \hline \hline
70 & 70 & 70 & 70     & 70     & 70   & Steel          & 6.15        \\
80 & 80 & 80 & 80     & 80     & 80   & Steel          & 6.99        \\
90  & 90  & 90  & 90      & 90      & 90    & Steel          & 7.83        \\
100 & 100 & 100 & 100     & 100     & 100   & Steel          & 8.67        \\
100 & 100 & 100 & 100     & 100     & 100   & Fairy          & 8.65        \\
100 & 100 & 100 & 100     & 100     & 100   & Normal           & 8.58        \\
100 & 100 & 100 & 100     & 100     & 100   & Ice           & 8.54        \\
100 & 100 & 100 & 100     & 100     & 100   & Bug            & 8.52        \\
100 & 134 & 110 & 70      & 84      & 72    & Rock, Electric      & 6.28       \\ 
55  & 55  & 55  & 135     & 135     & 135   & Fairy, Ghost   & 9.84        \\ \hline
\end{tabular}
}
\caption{Estimated strength of imaginary Pok\'{e}mons or real Pok\'{e}mons not present in the dataset. The last one is the Pok\'{e}mon most frequently used in the official online single battles of the latest season at the time of the experiment, yielding higher rating than others with the same or higher total stat.} \label{poke_hyp}
\end{center}
\end{table*}

\section{\uppercase{Discussion}}
In asymmetric environments, the advantage adjuster will help us to understand the strength of the unfairness of the environment, since NBTR learns it simultaneously along with quantification of the property. Using a small linear layer for $A$ as we did in our experiment should make it easily explainable.

Feature importance explanation methods such as DeepSHAP \cite{lundberg2017unified} can be useful to understand what feature matters more for strength when used to a NBTR estimator. Unlike applying it on a normal NN predictor of the winner, it will prevent the feature relates to the intransitivities like rock-paper-scissors mechanics from getting high importance since their influence should be eliminated from rating estimator's scalar output. In other words, feature importance on a normal classifier will tell you what decides the winner of the match, while the one on a NBTR estimator will tell you what decides overall strength.

\section{\uppercase{Conclusion}}
In this paper, we proposed NBTR, an ML framework to quantify properties and estimate those values of unknown items by integrating the Bradley-Terry model into neural network structures. Our method successfully quantified desired properties in both symmetric and asymmetric experimental settings.

Our framework provides a new ground for data mining and poses an alternative to the format of dataset, especially in environments where comparison data is structurally easier to collect. In online platforms where people choose and click something, we can just record those choices as training data. The same can be said for online multi-player games, as it creates a lot of match results between decks or parties. Even when a survey is required, gathering data of comparisons can be a better option than graded human scores, since comparisons are more precise and reliable, and our rating based on it provides a strong insight into the outcome of comparisons between items. We expect a lot of such applications being conducted in the future.

\section*{\uppercase{Acknowledgements}}

I would like to thank Professor Hideki Tsuiki for meaningful discussions. I am also grateful to the ICAART referees for useful comments.

\bibliographystyle{apalike}
{\small
\bibliography{biblo}}

\begin{thebibliography}{}

\bibitem[Bradley and Terry, 1952]{bradley1952rank}
Bradley, R.~A. and Terry, M.~E. (1952).
\newblock Rank analysis of incomplete block designs: I. the method of paired
  comparisons.
\newblock {\em Biometrika}, 39(3/4):324--345.

\bibitem[Elo and Sloan, 1978]{elo1978rating}
Elo, A.~E. and Sloan, S. (1978).
\newblock The rating of chessplayers: Past and present.

\bibitem[Ford, 1957]{doi:10.1080/00029890.1957.11989117}
Ford, L. R.~J. (1957).
\newblock Solution of a ranking problem from binary comparisons.
\newblock {\em The American Mathematical Monthly}, 64(8P2):28--33.

\bibitem[He et~al., 2016]{he2016deep}
He, K., Zhang, X., Ren, S., and Sun, J. (2016).
\newblock Deep residual learning for image recognition.
\newblock In {\em Proceedings of the IEEE conference on computer vision and
  pattern recognition}, pages 770--778.

\bibitem[Herbrich et~al., 2006]{herbrich2006trueskill}
Herbrich, R., Minka, T., and Graepel, T. (2006).
\newblock Trueskill^^e2^^84^^a2: a bayesian skill rating system.
\newblock {\em Advances in neural information processing systems}, 19.

\bibitem[Hinton and Salakhutdinov, 2006]{hinton2006reducing}
Hinton, G.~E. and Salakhutdinov, R.~R. (2006).
\newblock Reducing the dimensionality of data with neural networks.
\newblock {\em science}, 313(5786):504--507.

\bibitem[Hunter, 2004]{hunter2004mm}
Hunter, D.~R. (2004).
\newblock Mm algorithms for generalized bradley-terry models.
\newblock {\em The annals of statistics}, 32(1):384--406.

\bibitem[Inan et~al., 2016]{inan2016tying}
Inan, H., Khosravi, K., and Socher, R. (2016).
\newblock Tying word vectors and word classifiers: A loss framework for
  language modeling.
\newblock {\em arXiv preprint arXiv:1611.01462}.

\bibitem[Jalili et~al., 2018]{Jalili2018EvaluatingCF}
Jalili, M., Ahmadian, S., Izadi, M., Moradi, P., and Salehi, M. (2018).
\newblock Evaluating collaborative filtering recommender algorithms: A survey.
\newblock {\em IEEE Access}, 6:74003--74024.

\bibitem[Kingma and Ba, 2014]{kingma2014adam}
Kingma, D.~P. and Ba, J. (2014).
\newblock Adam: A method for stochastic optimization.
\newblock {\em arXiv preprint arXiv:1412.6980}.

\bibitem[LeCun et~al., 1998]{lecun1998gradient}
LeCun, Y., Bottou, L., Bengio, Y., and Haffner, P. (1998).
\newblock Gradient-based learning applied to document recognition.
\newblock {\em Proceedings of the IEEE}, 86(11):2278--2324.

\bibitem[Li et~al., 2021]{li2021neural}
Li, S., Ma, H., and Hu, X. (2021).
\newblock Neural image beauty predictor based on bradley-terry model.
\newblock {\em arXiv preprint arXiv:2111.10127}.

\bibitem[Lundberg and Lee, 2017]{lundberg2017unified}
Lundberg, S.~M. and Lee, S.-I. (2017).
\newblock A unified approach to interpreting model predictions.
\newblock {\em Advances in neural information processing systems}, 30.

\bibitem[Maystre, 2015]{choix}
Maystre, L. (2015).
\newblock choix.
\newblock https://choix.lum.li/en/latest/index.html.

\bibitem[Menke and Martinez, 2008]{menke2008bradley}
Menke, J.~E. and Martinez, T.~R. (2008).
\newblock A bradley--terry artificial neural network model for individual
  ratings in group competitions.
\newblock {\em Neural computing and Applications}, 17:175--186.

\bibitem[T7, 2017]{pokemon}
T7 (2017).
\newblock Pokemon- weedle's cave.
\newblock https://www.kaggle.com/datasets/terminus7/pokemon-challenge.

\bibitem[Xi et~al., 2022]{9446537}
Xi, W.-D., Huang, L., Wang, C.-D., Zheng, Y.-Y., and Lai, J.-H. (2022).
\newblock Deep rating and review neural network for item recommendation.
\newblock {\em IEEE Transactions on Neural Networks and Learning Systems},
  33(11):6726--6736.

\bibitem[Zhao et~al., 2020]{gorating}
Zhao, R., Dang, R., and Zhao, Y. (2020).
\newblock A neural network go rating model considering winning rate.
\newblock In {\em Proceedings of the 2019 3rd International Conference on
  Computer Science and Artificial Intelligence}, CSAI '19, page
  23^^e2^^80^^9327, New York, NY, USA. Association for Computing Machinery.

\end{thebibliography}

\end{document}